%% file: main.tex
\documentclass{article} % For LaTeX2e
\usepackage{mics/iclr2026_conference,times}

\input{mics/math_commands.tex}

\usepackage{hyperref}
\usepackage{url}

\definecolor{custom_velvet}{RGB}{148,0,64}
\definecolor{custom_green}{RGB}{0,148,64}
\hypersetup{
    colorlinks=true,
    citecolor=custom_velvet,
    linkcolor=custom_green,
    urlcolor=custom_velvet
}

\usepackage{multirow}
\usepackage[table]{xcolor}
\usepackage{enumitem} 
\usepackage{wrapfig}
\usepackage{sidecap}
\usepackage{adjustbox}
\usepackage{float}

\usepackage{microtype}
\usepackage{graphicx}
\usepackage{subfigure}
\usepackage{booktabs}

\usepackage{amsmath}
\usepackage{amssymb}
\usepackage{mathtools}
\usepackage{amsthm}
\usepackage{stfloats}

\makeatletter
\usepackage{xspace}
\def\@onedot{\ifx\@let@token.\else.\null\fi\xspace}
\DeclareRobustCommand\onedot{\futurelet\@let@token\@onedot}

\newcommand{\eqnref}[1]{Eq\onedot~\ref{#1}}
\renewcommand{\figref}[1]{Fig\onedot~\ref{#1}}

\renewcommand{\secref}[1]{Section~\ref{#1}}
\newcommand{\tabref}[1]{Tab\onedot~\ref{#1}}
\newcommand{\appref}[1]{Appendix~\ref{#1}}

\def\eg{\emph{e.g}\onedot}
\def\ie{\emph{i.e}\onedot}

\title{No Alignment Needed for Generation: Learning Linearly Separable Representations in Diffusion Models}

\author{Junno Yun, Ya\c{s}ar Utku Al\c{c}alar \& Mehmet Ak\c{c}akaya \\
Department of Electrical \& Computer Engineering, University of Minnesota, MN, USA \\
\texttt{\{yun00049,alcal029,akcakaya\}@umn.edu} \\
}
\iclrfinalcopy % Uncomment for camera-ready version, but NOT for submission.
\begin{document}

\maketitle

\begin{abstract}
Efficient training strategies for large-scale diffusion models have recently emphasized the importance of improving discriminative feature representations in these models. 
A central line of work in this direction is representation alignment with features obtained from powerful external encoders, which improves the representation quality as assessed through \emph{linear probing}. Alignment-based approaches show promise but depend on large pretrained encoders, which are computationally expensive to obtain. In this work, we propose an alternative regularization for training, based on promoting the \textbf{L}inear \textbf{SEP}arability (LSEP) of intermediate layer representations. LSEP eliminates the need for an auxiliary encoder and representation alignment, while incorporating linear probing directly into the network’s learning dynamics rather than treating it as a simple post-hoc evaluation tool. Our results demonstrate substantial improvements in both training efficiency and generation quality on flow-based transformer architectures such as SiTs, achieving an FID of 1.46 on $256 \times 256$ ImageNet dataset.
\end{abstract}

\begin{figure*}[ht]
    \centering
    \includegraphics[width=1.0\linewidth]{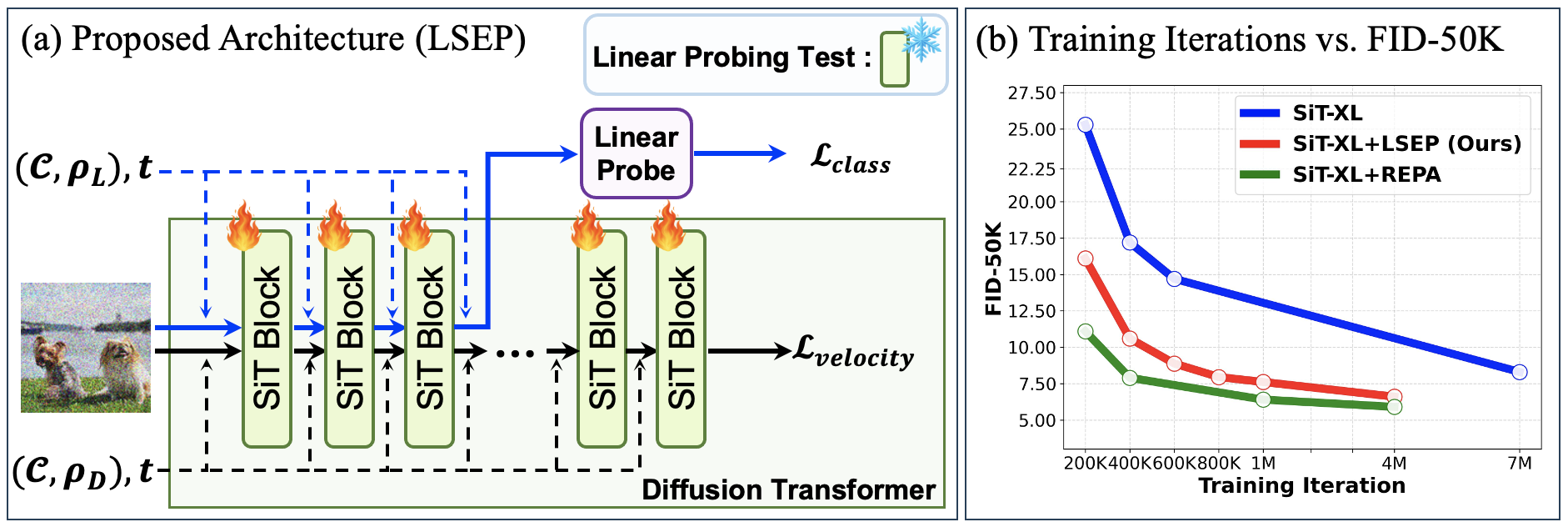} 
    \vspace{-0.4cm}
    \caption{(a) Overview of the proposed \emph{Linear SEParability (LSEP)} regularization. Unlike the linear probing test~\citep{alain2016linearprobe}, where separability of a target layer is evaluated on a frozen model, LSEP unlocks the layers and jointly optimizes a linear probe branch/classifier along with the denoising process. This actively drives the target layers toward \emph{higher linear separability} of features, substantially enhancing the generative model’s effectiveness on the denoising task \emph{without relying on any large-scale external encoder}. (b) SiT-XL with LSEP exhibits markedly faster FID improvement (without classifier-free guidance) than baseline SiT-XL, and by 4M iterations, its performance is comparable to that of the alignment-based model (REPA).}
    \label{fig:overview}
\end{figure*}

\section{Introduction} \label{sec:Introduction}
Diffusion models have demonstrated remarkable generative quality in a wide range of visual tasks, motivating additional architectural innovations to  further improve their performance~\citep{dhariwal2021beatGANs, karras2022elucidating, peebles2023scalable, ma2024sit}. These models inherently learn semantically meaningful representations through the noise-prediction objective used in denoising~\citep{baranchuk2022labelefficient, xiang2023DDAE, chen2025deconstructing}. However, their learned representations are not explicitly optimized for representation learning and are often less expressive than models trained specifically for this purpose~\citep{xiang2023DDAE, yu2025REPA}. 

From a representational perspective, a growing body of research has examined how improving learned representations can benefit diffusion training~\citep{zhu2024sd, yu2025REPA, jiang2025SREPA, leng2025REPAE, yao2025reconstruction}. These studies have explored methods to improve the training efficiency and generative performance of transformer-based diffusion models through \textit{representation alignment}. In particular, they have consistently shown that acquiring strong representations at specific network depths by aligning with high-quality internal~\citep{zhu2024sd, jiang2025SREPA} or external~\citep{yu2025REPA, leng2025REPAE, yao2025reconstruction} representations significantly accelerates convergence and enhances the quality of generated outputs.

A recent powerful approach, representation alignment (REPA)~\citep{yu2025REPA}, leverages high-quality representations extracted from large-scale pre-trained transformer models, such as DINOv2~\citep{oquab2024dinov} and CLIP~\citep{radford2021CLIP}. REPA explicitly aligns early-stage diffusion transformer features with clean image features from external encoders, thereby encouraging stronger representational capacity. This alignment enables deeper layers to concentrate on high-frequency content, ultimately improving generative performance. However, training such pre-trained visual encoders demands access to large-scale datasets and involves substantial computational costs.

An alternative approach leverages self-representation alignment~\citep{zhu2024sd, jiang2025SREPA} without external encoders. In this framework, self-representation alignment (SRA)~\citep{jiang2025SREPA} adapts teacher–student discriminative pair structures within diffusion models to enable self-supervised knowledge distillation~\citep{zhu2024sd}. Using the observation that deeper layers produce richer representations~\citep{xiang2023DDAE, yu2025REPA}, the teacher is assigned to deeper layers with lower noise, while the student is associated with the earlier layers. This strategy has been shown to enhance training effectiveness without relying on pre-trained encoders. Nevertheless, the approach remains fundamentally constrained by the representational capacity of the diffusion model itself. 

To assess the effectiveness of learned representations, prior work commonly employs \textit{linear probing}~\citep{alain2016linearprobe}. In this evaluation protocol, the trained diffusion model is frozen, and features from specific layers are extracted to train a lightweight linear probe (classifier). The classifier’s accuracy serves as a measure of the \emph{degree of linear separability} of the feature space. Linear probing accuracy has been shown to correlate strongly with both the training efficiency and the generative quality of diffusion transformers~\citep{leng2025REPAE, yu2025REPA}. These findings suggest that a key factor underlying the success of prior approaches is the refinement of feature representations, thereby promoting \emph{linear separability}.

\vspace{-6pt}
\paragraph{Our Approach:} This work is motivated by a simple yet fundamental question:

{\centering
\emph{“Can diffusion models learn highly linearly separable representations that improve training efficiency while producing higher-quality outputs without representation alignment or external encoders?”}\par}

%Since \emph{linear separability} reflects the quality of diffusion model training, 
To this end, 
we introduce \textbf{L}inear \textbf{SEP}arability (LSEP), a training regularization strategy that incorporates a linear probe into diffusion models to simultaneously improve the separability of early-stage hidden representations and optimize the denoising objective. 

In particular, we insert a trainable linear probe into an intermediate layer of the diffusion model, as shown in ~\figref{fig:overview}, similar in spirit to linear probing evaluations, but \emph{without freezing} the model parameters. Unlike prior alignment-based approaches, our method does \emph{not require an external visual encoder or explicit representation alignment}, while still promoting stronger linear separability at the intended depth of the diffusion model. Experiments show that LSEP substantially boosts both the training efficiency and the generative output quality %quality of generated outputs 
on flow-based transformer SiTs~\citep{ma2024sit}. 

Our main contributions are as follows:
\begin{itemize}[leftmargin=*, itemsep=0em, topsep=0pt]

    \item We introduce \emph{Linear SEParability (LSEP)}, a framework that integrates a linear probe (classifier) into generative model training to simultaneously enhance linear separability and the standard denoising objective. 
    
    \item To enable the two objectives to mutually reinforce each other, we propose novel training techniques for the classification term while keeping the denoising training unchanged: (1) classification-specific conditioning for the linear probe branch, (2) random cropping to enhance patch-level linear separability, and (3) time-dependent weighting of the classification loss. 
    
    \item We demonstrate that LSEP substantially enhances both training efficiency and generative quality in flow-based transformer architectures, without relying on \emph{an external visual encoder or explicit representation alignment}.
        
    \item SiT-XL with LSEP converges to lower FID significantly faster than the baseline SiT-XL, reaching performance comparable to the alignment-based model REPA, as shown in~\figref{fig:overview} (b). Moreover, it achieves an FID of 1.46 with classifier-free guidance using the guidance interval, establishing the best performance among models without relying on external encoder architectures.

    \item Finally, we show that LSEP synergistically combines with alignment-based methods to enhance linear separability of representations through distinct mechanisms, and to further improve both training efficiency and generative performance.   

\end{itemize}

\section{Preliminaries}

\subsection{Training Flow-based Diffusion Transformer} \label{sec:sec2_1}
Flow-based approaches~\citep{lipman2023flow} learn a velocity field \( \mathbf{v}_\theta(\mathbf{x}_t, t) \) that defines a probability flow ordinary differential equation (PF-ODE), characterizing the deterministic evolution of a data point \( \mathbf{x}_t \). SiT~\citep{ma2024sit} adopts this framework to model a continuous-time forward process: 
\begin{equation}
    \mathbf{x}_t = \alpha_t \mathbf{x}_0 + \sigma_t \boldsymbol{\epsilon}, \quad \boldsymbol{\epsilon} \sim \mathcal{N}(\mathbf{0}, \mathbf{I}),
\end{equation}
where \( \alpha_t \) and \( \sigma_t \) are time-dependent functions such that \( \alpha_t \) decreases and \( \sigma_t \) increases with \( t \in [0, T] \), satisfying \( \alpha_0 = \sigma_T = 1 \) and \( \alpha_T = \sigma_0 = 0 \). The PF-ODE is given by
\(
\frac{d\mathbf{x}_t}{dt} = \mathbf{v}_\theta(\mathbf{x}_t, t),
\)
where the distribution of the ODE solution at time \( t \) matches the marginal distribution of the forward process. 
To train the velocity model \( \mathbf{v}_\theta(\mathbf{x}_t, t) \), the following velocity matching loss is minimized:
\begin{equation}
    \mathcal{L}_{\text{velocity}} = \mathbb{E}_{\mathbf{x}_0, \boldsymbol{\epsilon}, t} \left[ \left\| \mathbf{v}_\theta(\mathbf{x}_t, t) - \dot{\alpha}_t \mathbf{x}_0 - \dot{\sigma}_t \boldsymbol{\epsilon} \right\|^2 \right],
\label{eq:loss_velocity}
\end{equation}
where \( \dot{\alpha}_t = \frac{d\alpha_t}{dt} \) and \( \dot{\sigma}_t = \frac{d\sigma_t}{dt} \) denote the time derivatives of \( \alpha_t \) and \( \sigma_t \), respectively.

In practice, to incorporate class guidance via classifier-free guidance (CFG)~\citep{ho2021classifierfree}, the velocity model \( \mathbf{v}_\theta \) is trained conditionally on a label $ \mathbf{c} \in \mathcal{C} $, resulting in a model of the form \( \mathbf{v}_\theta(\mathbf{x}_t, t \mid \mathbf{c}) \)~\citep{peebles2023scalable, ma2024sit, yu2025REPA}. Here, $\mathcal{C}$ includes both the dataset class labels $\{\mathbf{c}_{\text{class}}\}$
and an unconditional label $\mathbf{c}_{\varnothing}$, \ie $\mathcal{C} = \{\mathbf{c}_{\text{class}}\} \cup \mathbf{c}_{\varnothing}$. Both types of conditions are employed during training, where the unconditional label is used with a small probability, $\rho_D \ll 1.$

\subsection{Representation Alignment for Generative Models} \label{sec:repa}
To improve training efficiency and generation quality of diffusion transformers, REPA~\citep{yu2025REPA} aligns a model’s hidden-layer representations with large-scale pretrained self-supervised visual encoders such as DINOv2~\citep{oquab2024dinov} and CLIP~\citep{radford2021CLIP}. Specifically, REPA introduces a regularization term that maximizes patch-wise similarity between features $\mathbf{y}_{\mathrm{clean}}$, extracted from the pretrained external encoder, and $k^{\text{th}}$ hidden layer features $\mathbf{h}_t^k$ of the diffusion transformer encoder at timestep $t$. The representation alignment loss is defined as:
\begin{equation}
\mathcal{L}_{\textrm{repa}} = 
- \mathbb{E}_{\mathbf{x}_{\textrm{clean}}, \epsilon ,t} \left[ 
    \frac{1}{N} \sum_{n=1}^{N} \textrm{sim}_\textrm{cos}\!\left(\mathbf{y}_{\textrm{clean}}^{n}, \, \textrm{MLP}(\mathbf{h}_t^{k, n})\right)
\right],
\label{eq:loss_repa}
\end{equation}
where $N$ is the total number of patches, $\epsilon$ is the noise level at timestep $t$, and $\text{MLP}$ is a trainable multilayer perceptron that projects $\mathbf{h}_t^{k, n}$ to adaptively align it with the representation space of $\mathbf{y}_{\textrm{clean}}^n$, while $\textrm{sim}_{\textrm{cos}}(\cdot, \cdot)$ denotes the cosine similarity function. This regularization term is incorporated into the original diffusion-based objective in~\eqnref{eq:loss_velocity} during training, yielding the combined loss:
\begin{equation}
\mathcal{L}_{\text{REPA}} = \mathcal{L}_{\text{velocity}} + \lambda \cdot \mathcal{L}_{\text{repa}}.
\label{eq:loss_repa_total}
\end{equation}
Large-scale pretrained encoders (\eg, DINOv2) are optimized not only with image-level but also with patch-level objectives, enabling them to learn more robust representations~\citep{zhou2022image, oquab2024dinov}. From the perspective of the patch-level manifold space~\citep{hao2022learning}, REPA leverages patch-wise representation alignment to capture these finer-grained structures, thereby maximizing the effectiveness of alignment with such pretrained encoders.

\subsection{Linear Probing}
Linear probing evaluations~\citep{alain2016linearprobe} were originally proposed to analyze deep neural networks by measuring the degree of linear separability of features across layers. In this approach, the model under investigation is frozen. Given an input $\mathbf{x}$, the neural network encoder extracts features $\mathbf{h}^k$ at the $k^{\text{th}}$ layer, analogous to~\secref{sec:repa}, which are subsequently reduced via global pooling to yield the representation $\mathbf{g}^k$. A linear classifier $f^k$ is then trained on these pooled features to solve the classification task by minimizing the cross-entropy loss:
\begin{equation}
\mathcal{L}_\text{class} 
= - \mathbb{E}_{(\mathbf{x}, \mathbf{c})} 
\big[ \, \mathbf{c}_{\textrm{gt}}^{\top} \log \big( f^{k}(\mathbf{g}^{k}) \big) \big],
\label{eq:loss_class}
\end{equation}
where $\mathbf{c}_{\textrm{gt}}$ is ground-truth label and $f^k(\mathbf{g}^{k}) = \text{softmax}(\mathbf{W} \mathbf{g}^{k} + \mathbf{b})$ with $(\mathbf{W}, \mathbf{b})$ as probe parameters. 

\looseness=-1
Recent studies have demonstrated that both transformer-based and CNN-based diffusion models learn semantically meaningful representations through the denoising task in training~\citep{baranchuk2022labelefficient, chen2025deconstructing, xiang2023DDAE}, as evidenced by evaluations of their linear separability via linear probing. Furthermore, alignment-based methods have employed linear probing to quantify the improvements in the linear separability of the target layers’ representations, demonstrating how such enhancements correlate with training efficiency~\citep{jiang2025SREPA, yu2025REPA}.

\section{LSEP: Promoting Linear Separability of Denoising Networks} \label{sec:LSEP}

We introduce a simple yet novel trainable linear probe into an intermediate layer of the diffusion model, in the spirit of linear probing evaluations, but train it \emph{without freezing} its model parameters. We conceptualize our proposed method as a harmonious integration of diffusion and linear classification tasks, designed to enhance both simultaneously, as described in~\figref{fig:overview}. All intermediate layers, along with a linear probe inserted at a pre-specified depth, are simultaneously trained to enhance linear separability of the intermediate representations. 
The entire diffusion model leverages these well-separated features to perform effective denoising. 

Consequently, our model should be carefully designed to accommodate the coexistence of these two distinct tasks. To this end, we introduce several novel training strategies: (1) design of the linear probe branch, (2) task-specific conditioning of the shared intermediate layers, (3) random cropping for improving patch-level linear separability, and (4) time-dependent weighting scheduling of the linear probing loss. Each subsequent subsection provides details on these components, and their respective results are presented in~\secref{sec:Key_Strategies}.

\subsection{Linear Classifier Built on Diffusion Model} \label{sec:method_linear_probe}
\begin{wrapfigure}{r}{0.45\textwidth} 
    \vspace{-0.8cm}
    \centering
    \includegraphics[width=\linewidth]{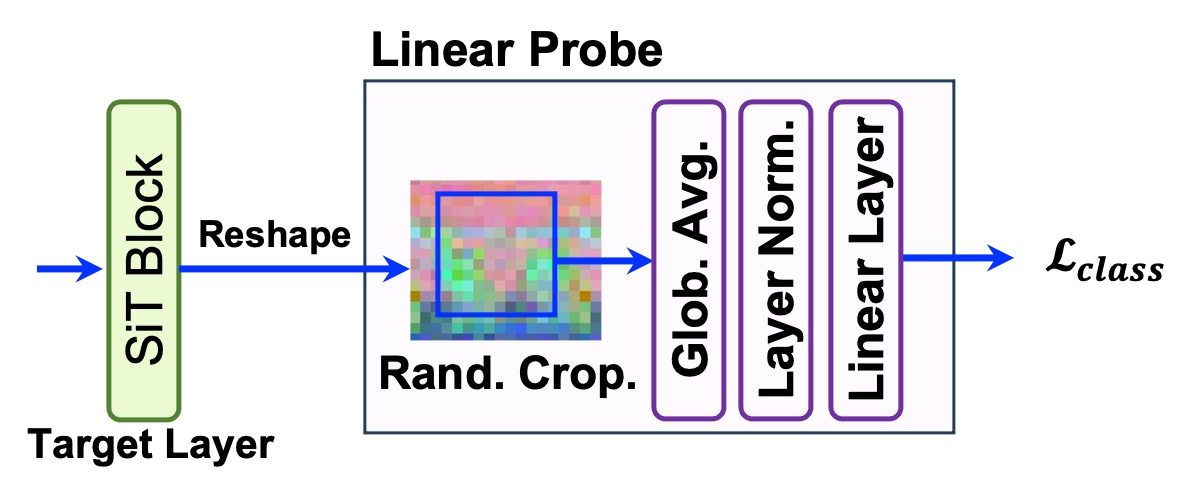}
    \vspace{-0.7cm}
    \caption{Architectures of the linear classifier.}
    \label{fig:linear_probe}
    \vspace{-0.3cm}
\end{wrapfigure}    
\looseness=-1
As illustrated in~\figref{fig:linear_probe}, our linear classifier consists of a normalization module followed by a linear layer, similar in spirit to prior works~\citep{alain2016linearprobe, he2022masked, chen2025deconstructing, yu2025REPA}. Unlike these works, which may or may not employ batch normalization, we adopt layer normalization to ensure stable training in the presence of the denoising objective. Thus, intermediate features extracted from the SiT model are globally aggregated via average pooling for dimensionality reduction, followed by layer normalization and a linear classification head. The diffusion transformer is trained up to a specified target depth jointly with the classifier. This joint optimization encourages early-stage representations to become more linearly separable, guided by supervision from the classifier. The classification loss is given in~\eqnref{eq:loss_class}, with the only modification being the addition of layer normalization.

Our proposed total loss, which incorporates the scaling factor \( \omega_{\text{class}} \) for the classification term to control the trade-off between classification and denoising tasks, together with~\eqnref{eq:loss_velocity}, is defined as:
\begin{equation}
\mathcal{L}_{\text{LSEP}} = \mathcal{L}_{\text{velocity}} + \omega_{\text{class}} \cdot \mathcal{L}_{\text{class}}.
\end{equation}

\subsection{Conditioning to Linear Probe Branch}  \label{sec:32}
In practice, conditioning of the diffusion transformer consists of time and class embeddings, which are summed and applied through projection operators. For time embedding, we use the same timestep conditioning for both tasks, which helps the classifier to align with the temporal dynamics of the denoising process. 
However, we apply a different class conditioning to the linear probe branch, as indicated by the blue path in~\figref{fig:overview} to enhance linear separability of intermediate features. This is because the class embedding is a function of the class label, $\mathbf{c}_\textrm{class}$. Thus, this information may cause shortcut learning, where the linear classifier relies directly on class conditioning to perform classification rather than learning linearly separable representations. 

To mitigate this, we assign the unconditional class label ($\mathbf{c}_{\varnothing}$) as $\mathbf{c}_\textrm{class}$ with probability $\rho_L$, chosen to be close to 1. This withholds the class information from the classifier and prevents it from relying solely on class conditioning. With probability $1-\rho_L$, we assign another class label ${\bf c} \in {\cal C} - \{\mathbf{c}_{\varnothing}\}$ as $\mathbf{c}_\textrm{class}$, which exposes the classifier to non–affine-transformed features. Note this approach is different from the class conditioning used for the denoising task of the diffusion model, where non-null class information is included in the class conditioning with higher probability, $1-\rho_D$. Our combined strategy encourages feature representations in the early stages to capture meaningful semantics.

\subsection{Random Cropping for Improving Patch-level Linear Separability} 
\begin{figure*}[ht]
    \centering
    \vspace{-0.2cm}
    \includegraphics[width=0.9\linewidth]{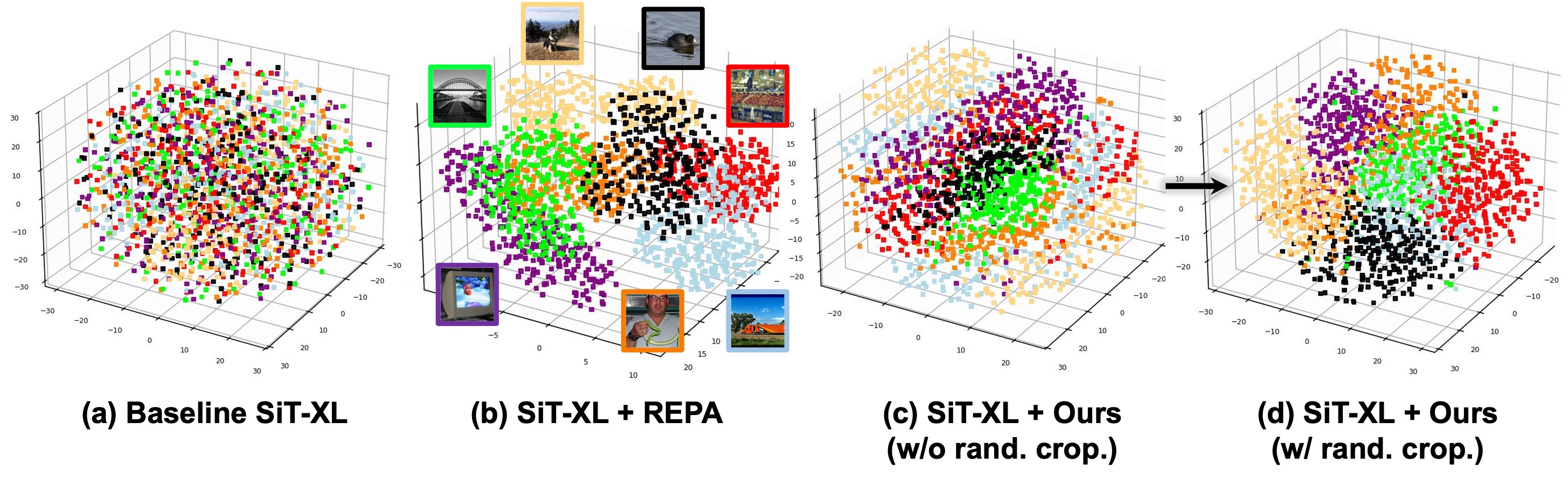} 
    \vspace{-0.4cm}
    \caption{t-SNE 3D projection visualization of the patch-level manifold space of SiT-XL using baseline, REPA, and our method with and without random cropping (400K iters, $t$=0.7). Each patch within the $8^{th}$ layer intermediate features is represented by a square, with distinct colors for 7 classes. 
    }
    \label{fig:3d_patch_level}
\end{figure*}

Our linear probe uses a summary statistic of the intermediate features, obtained via global pooling, to perform its classification task. While this ensures linear separability of the pooled feature vector, the patch-level features may not be as optimally separated in the patch-level manifold space~\citep{hao2022learning}. To further investigate this, we performed a t-SNE~\citep{maaten2008visualizing} 3D visualization in the patch-level manifold space, as illustrated in~\figref{fig:3d_patch_level}. In~\figref{fig:3d_patch_level} (b), the alignment-based method (REPA) forms well-separated clusters by leveraging patch-wise alignment with an extensively pre-trained external encoder. Our method (\figref{fig:3d_patch_level} (c)), optimized with globally pooled features, achieves improved clustering compared to the baseline (a). However, certain patches (in light orange and light blue) exhibit a dispersed distribution.

To address this, we randomly crop the intermediate feature map into $n \times n$ patches and compute the mean over this subset before feeding it to the linear classifier as illustrated in~\figref{fig:linear_probe}. Concretely, we first reshape the intermediate features from $\mathbf{h}^k \in \mathbb{R}^{T \times D}$ to $\mathbb{R}^{t \times t \times D}$, where $T$ is the length of features, $D$ is the channel dimension, and $t^2 = T$. We then randomly crop the features to $\mathbb{R}^{n \times n \times D}$, with $n \le t$. This strategy enhances the separability of both the mean vectors and subsets of patch-level features, while also introducing diversity similar to data augmentation. As illustrated in~\figref{fig:3d_patch_level} (d), this promotes clearer cluster formation and further contributes to improved denoising training.

\subsection{Time-Dependent Weighting Scheduling of $ \mathcal{L}_{class} $}
\looseness=-1
One of the major differences between standard classification and classification within our designed diffusion model is that the inputs are combined with different noise levels, resulting in a multitude of input distributions. Although the time embedding effectively tracks these variations and guides the corresponding denoising tasks, a single linear probe has limitations in classifying across such diverse distributions. While incorporating a time-dependent embedding into the classifier or using multiple classifiers for different time steps may be viable, our focus is not on enhancing the classifier head. Instead, we maintain a single classifier and introduce diversity through weight scheduling to improve the linear separability of intermediate transformer blocks.

To this end, we apply a time-dependent piecewise constant weighting scheme to the linear probe loss, allowing the linear probe to assign different weights according to the noise level and thus learn more effectively. This approach divides the different noise level distributions into \(k\) groups, allowing the linear probe to be optimized for each group. It also assigns larger weights to higher noise levels, which helps the classifier perform more effectively on noisier inputs. The time-dependent weight, \(\omega_{\text{class}}(t)\) with \(k\) bins is defined as:
\begin{equation}
\omega_{\text{class}}(t, k) =
\omega_{\text{start}} + \lfloor t \cdot k \rfloor \cdot \Delta \omega, \quad t \in [0,1]
\end{equation}
where $\Delta \omega = (\omega_{\text{max}} - \omega_{\text{min}})/k$, and $\omega_{\text{min}}$ and $\omega_{\text{max}}$ are the minimum and maximum weighting values, respectively. 
We denote this as 
\(\omega_{\text{class}}(t, k) = [\omega_{\text{start}}, \omega_{\text{end}}]_{k\text{ bins}}\) 
for future reference.

\input{Tables/result_table1}

\section{Experimental Setup} 

\subsection{Implementation details}
We closely follow the experimental setup of SiT~\citep{ma2024sit} and REPA~\citep{yu2025REPA}, unless stated otherwise. All models are trained and evaluated on ImageNet~\citep{deng2009imagenet}, where images are preprocessed to a resolution of 256$\times$256 following the data preprocessing protocol of ADM~\citep{dhariwal2021beatGANs}. We employ the Base, Large, and X-Large model variants introduced in SiT~\citep{ma2024sit}, all configured with a patch size of 2. We insert a linear classifier after the 4\textsuperscript{th}, 7\textsuperscript{th}, and 8\textsuperscript{th} layers in the SiT-B/2, SiT-L/2, and SiT-XL/2 models, respectively.

To ensure fair comparison with prior work~\citep{ma2024sit, yu2025REPA}, we train all models using a consistent global batch size of 256. Optimization is performed using AdamW~\citep{kinga2015method, loshchilov2018decoupled} with a constant learning rate of $1\times10^{-4}$ for training the diffusion model. We provide the detailed hyperparameters in~\appref{appdx:implementation_details}. 

\subsection{Evaluation Protocol}
For evaluating image generation quality, we strictly follow the ADM evaluation protocol~\citep{dhariwal2021beatGANs}. We report several standard metrics, including Fréchet Inception Distance (FID)~\citep{heusel2017gans}, Structural FID (sFID)~\citep{nash2021generating}, Inception Score (IS)~\citep{salimans2016improved}, and Precision and Recall~\citep{kynkaanniemi2019improved}, all computed using 50K generated samples. Following the sampling procedures from SiT~\citep{ma2024sit} and REPA~\citep{yu2025REPA}, we adopt the SDE-based Euler–Maruyama sampler with 250 steps. All evaluations are performed on 50K validation images from the ImageNet dataset~\citep{deng2009imagenet}, resized to 256$\times$256 resolution.

\subsection{Analysis of Key Strategies} \label{sec:Key_Strategies} 

We present detailed experimental results for the SiT-L/2 model in~\tabref{tab:table_results}, effectively serving as an ablation study for the training innovations outlined in~\secref{sec:LSEP}.

\colorbox{blue!10}{\textbf{Class Conditioning for Linear Probe}} 
The unconditioning probability $\rho_L$ indicates the proportion of the null class $\mathbf{c}_{\varnothing}$, as in \secref{sec:32}. When the unconditioning probability is set to 0.1, matching that of the denoising model, the linear probe relies on the conditioning label for shortcut learning. This not only fails to improve the linear separability of intermediate features but also degrades generative performance. Conversely, setting it to $\rho_L=1.0$ (\ie, using only $\mathbf{c}_{\varnothing}$) prominently improves generative performance. However, this also causes the class conditioning for the classifier and the denoiser to be learned independently, leading to inconsistencies in class representation. Therefore, we find that using $\rho_L=0.9$ provides the optimal conditioning ratio. Further analysis is provided in~\appref{appdx:loss_class}.

\colorbox{orange!10}{\textbf{Target Depth}} Numerous works have demonstrated that efficient training of diffusion models depends on representation quality in early layers~\citep{yu2025REPA, jiang2025SREPA}. Our results align with these insights, showing that the optimal depth for incorporating the linear classifier corresponds to the shallow layers of the model. When the depth is shifted toward the middle layers (\eg, 9 or 10 for SiT-L/2), the generative performance degrades, as the reduced capacity for denoising limits overall effectiveness. Conversely, connecting to very early layers also hampers denoising training. We empirically find that layer 7 yields the best results for SiT-L, and adopt this configuration in our experiments. For SiT-B and SiT-XL, the optimal depths are found to be layers 4 and 8, respectively.

\colorbox{green!10}{\textbf{Random Cropping}} 
To enhance linear separability in the patch-level manifold space as shown in~\figref{fig:3d_patch_level}, we investigated the effects of random cropping with varying box sizes. We found that selecting a box size randomly from the range 12 to 16 yielded the best results. This introduces variability, providing more diverse feature samples for the linear classifier and improving separability not only of pooled features but also within the patch-level manifold space. 

\colorbox{cyan!10}{\textbf{Time-dependent $\omega_{\text{class}}$}} 
The first three rows compare a time-dependent \(\omega_{\text{class}}\) to a constant \(\omega_{\text{class}}\) in the absence of random cropping. This demonstrates that assigning different weights to \(\mathcal{L}_{\text{class}}\) over time, using a piecewise constant schedule with multiple bins for training a linear classifier improves the generative results. The last three rows further examine different weight intervals and numbers of bins when time-dependent \(\omega_{\text{class}}\) is combined with random cropping approach. The results show \([0.0275, 0.0325]_{\text{10 Bins}}\) achieves the best performance, and is used for the remainder of the study.

\section{Results}
\subsection{Linear Separability}

\begin{figure*}[h]
    \centering
    \includegraphics[width=1.0\linewidth]{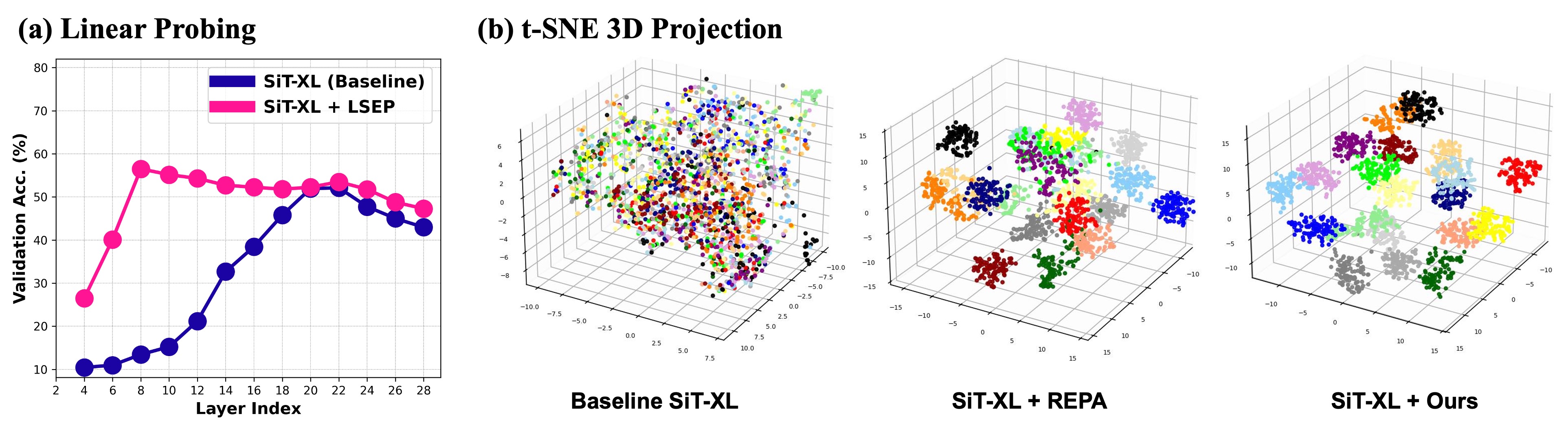} 
    \vspace{-0.5cm}
    \caption{(a) Linear probing evaluation results on baseline SiT-XL and SiT-XL+LSEP with $t=0.1$. (b) t-SNE 3D projections for baseline SiT-XL (left), alignment-based REPA (middle), and proposed method (right). Intermediate features are extracted from the $8^{\text{th}}$ layer of each pre-trained SiT-XL model with $t=0.1$. Each feature is globally pooled and sampled from 20 randomly selected classes, with 100 samples per class. The representations are then projected into 3D space, with each class visualized using a distinct color. For all methods, the 400K-iteration checkpoint was used.}
    \label{fig:linear_sep}
    \vspace{-0.2cm}
\end{figure*}

We evaluate the effect of promoting linear separability using linear probing, 3D projections, and PCA visualization for SiT-XL with our proposed method. \figref{fig:linear_sep} (a) shows that LSEP achieves higher overall linear probing accuracy across layers, particularly at the early stages. Unsurprisingly, linear probing performance is substantially improved using the LSEP training strategy that jointly targets a linear classification task. In~\figref{fig:linear_sep} (b), our proposed method produces well-defined clusters, comparable to those observed in REPA, which indicates that the classes are clearly linearly separable, even though no external encoder information is used. We also provide PCA visualizations in~\appref{appdx:pca_visual}, which demonstrate that LSEP preserves clearly separable components across varying noise levels.

\subsection{Quantitative Results} \label{sec:Quantitative Results}
\paragraph{Analysis Across Model Sizes.} 
\tabref{tab:table_results2} summarizes generative performance of models of different sizes with different training paradigms. Note the number of parameters for REPA \emph{excludes} the additional parameters related to the external encoder. Our method consistently improves the generation FID compared to the baseline SiT baseline trained for the same 400K iterations. Notably, FID of 12.3 (\tabref{tab:table_results2}) and IS of 98.3 (\tabref{tab:table_results}) achieved by our LSEP (SiT-L) outperform the 12.5 and 90.7, respectively, reported for REPA (SiT-L)~\citep{yu2025REPA} with pretrained external MAE-L encoder~\citep{he2022masked}, which contains an additional 304M parameters in addition to the transformer diffusion parameters.

\begin{wrapfigure}[14]{r}{0.5\textwidth}
\vspace{-1.3cm}
\begin{minipage}{0.5\textwidth}
    \input{Tables/result_table2}
\end{minipage}
\end{wrapfigure}

\begin{figure*}[b]
    \centering
    \includegraphics[width=1.0\linewidth]{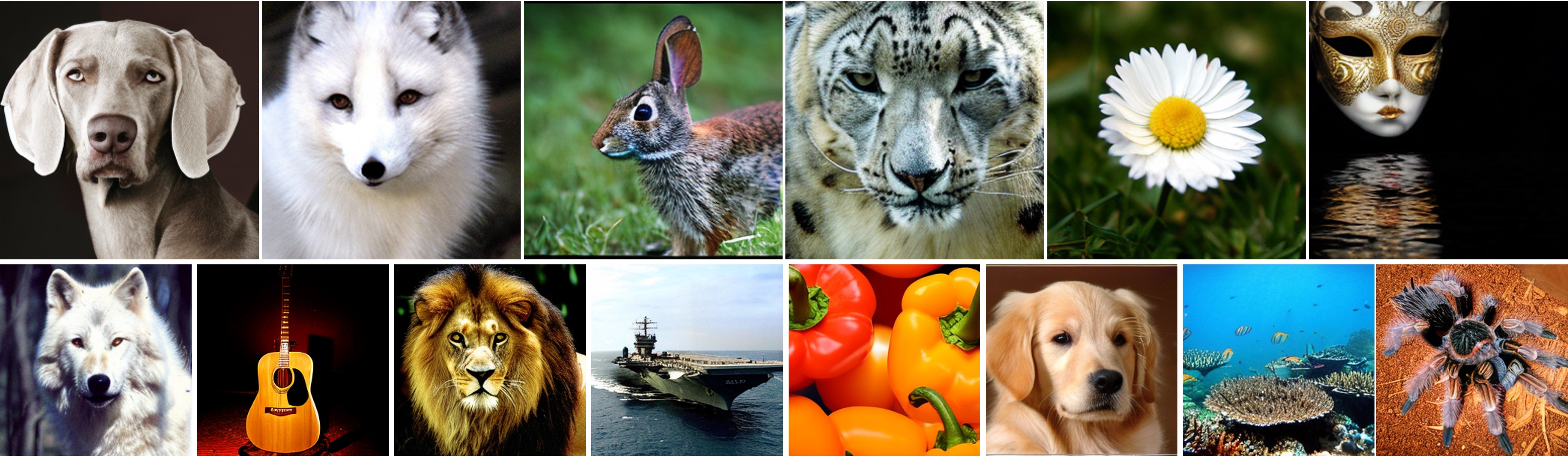} 
    \vspace{-0.5cm}
    \caption{Selected samples on ImageNet $256 \times 256$ from the SiT-XL/2 model with LSEP. We use classifier-free guidance with $\omega_{cfg} = 4.0$.}
    \label{fig:vis_results}
\end{figure*}

While LSEP shows substantial gains in early stages of trainings, we emphasize that LSEP continues to steadily improve FID throughout training, with consistent gains: by 4M iterations, its performance approaches that of REPA, while remaining well beyond the reach of baseline training at 7M iterations, as illustrated in~\figref{fig:overview} (b). Furthermore, our approach achieves these promising results \emph{without relying on any pre-trained external models or explicit representation alignment}. Selected qualitative results for SiT-XL/2 using our method are presented in~\figref{fig:vis_results}. More qualitative results are provided in~\appref{appdx:more_qual_results}.

\input{Tables/result_table3}

\paragraph{Combining Alignment-Based Methods with LSEP.}
REPA improves the representations by performing token-wise alignment, while our LSEP enhances linear separability by mainly utilizing mean representation vectors. Both methods individually improve linear separability; however, when used together, they are expected to provide more powerful representation geometry and improved training efficiency. As shown in~\tabref{tab:fid_comparison}, combining REPA with LSEP consistently drives FID even lower, proving that this synergy not only further accelerates training but also enhances generative quality. Additional analysis using patch-level 3D projection visualizations is presented in~\appref{appdx:repa_ours}.

\paragraph{System-Level Comparison.}

In~\tabref{tab:system_level}, we report evaluation results using CFG~\citep{ho2021classifierfree} with the guidance interval~\citep{kynknniemi2024applying} (please see~\appref{appdx:cfg_guidance} for details). Our method achieves a best FID of 1.46 among models that do not use an external visual encoder and is comparable to alignment-based models, such as REPA, which rely on pretrained external encoders. 

\section{Discussion}
\textbf{Limitations.} The scope of this paper does not cover advanced generative settings such as text-to-image generation, video generation, and 2K higher-resolution images. Further investigation is warranted to evaluate the applicability and scalability of our framework in these scenarios.

\textbf{Future Directions.}
Similar to how REPA variants~\citep{yao2025reconstruction,leng2025REPAE} improve generative performance by tuning the VAE through representation alignment, LSEP inherently offers an alternative means to achieve this. Furthermore, since most vision foundation models are transformer-based, prior studies on representation alignment have focused mainly on transformer diffusion models. In contrast, LSEP can also be applied to CNN-based models, broadening its applicability. These ideas will be investigated in future studies.

\section{Conclusion}

In this paper, we proposed LSEP, a regularization approach that independently enhances the linear separability of generative models without relying on large-scale pre-trained encoders or representation alignment. We demonstrated that linear separability is a core principle of diffusion training and repurposed linear probing, which is typically used for post-hoc evaluation, as an effective training tool. Our results show that LSEP improves both training efficiency and generative performance, achieving a state-of-the-art FID score of 1.46 with a single model, without the need for any alignment.

\bibliography{refs_utku_base}
\bibliographystyle{mics/iclr2026_conference}

\newpage
\appendix

%%%%%%%%%%%%%%%%%%%
%%%%% Appendix %%%%
%%%%%%%%%%%%%%%%%%% 

%%%%%%%%%%%%%%%%%%% Implementation details  %%%%%%%%%%%%%%%%%% 
\section{Implementation details} \label{appdx:implementation_details}
\input{Tables/table_hyperparameters}
Strictly following the experimental setups of SiT~\citep{ma2024sit} and REPA~\citep{yu2025REPA} for the denoising loss, we integrate the linear probe into our model and employ the hyperparameters listed in~\tabref{tab:table_hyperparameters}. We use an increased learning rate for the linear classifier in the SiT-B/2 model, which significantly improves both training efficiency and generative performance. On the other hand, for the Large and X-Large models, this adjustment does not lead to improvements in generative results. All experiments were conducted using 4 NVIDIA A100 GPUs.

\section{Additional Analysis}

%%%%%%%%%%%%%%%%%%  Conditioning for linear probe branch  %%%%%%%%%%%%%%%%%% 
\subsection{Conditioning for linear probe branch}  \label{appdx:loss_class}

\begin{wrapfigure}{r}{0.4\textwidth} 
    \vspace{-0.7cm}
    \centering
    \includegraphics[width=0.4\textwidth]{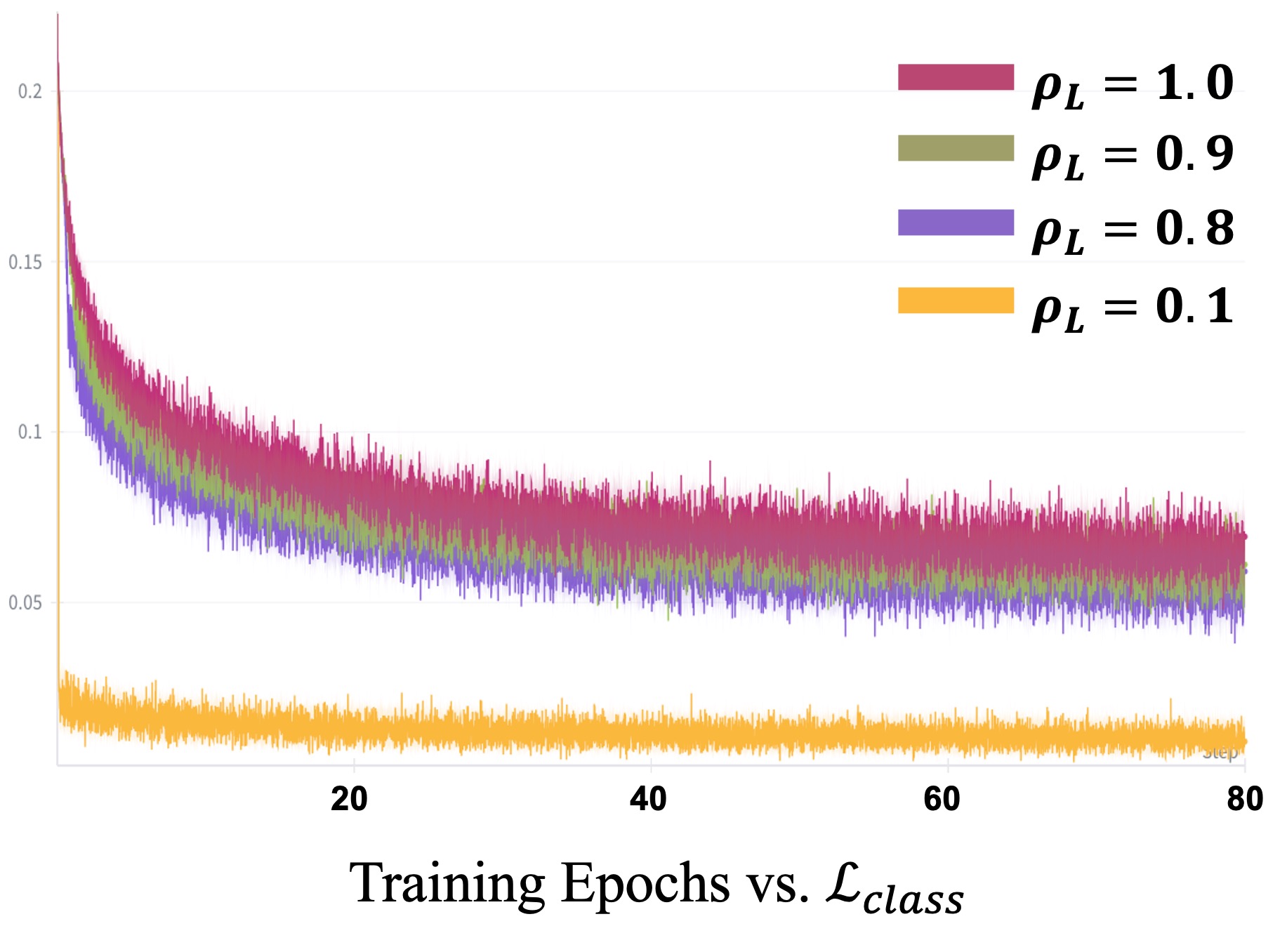} 
    \vspace{-0.5cm}
    \caption{Comparison of $\mathcal{L}_{\text{class}}$ across training epochs with varying conditioning ratios in the linear probe branch.}
    
    \label{fig:appdx_loss_class}
\end{wrapfigure}

As discussed in~\secref{sec:Key_Strategies}, conditioning for linear probing plays an important role in improving both linear separability and generation quality. When the unconditioning probability $\rho_L$ in~\secref{sec:32} is set to 0.1, matching that of the denoising model, the linear probe tend to rely on shortcut learning, as illustrated by the yellow curve in~\figref{fig:appdx_loss_class}. Moreover, as the unconditioning ratio increases, the classifier’s reliance on class information gradually diminishes, leading to a more progressive learning process. However, as mentioned earlier, when the conditioning is fully assigned to the unconditional class (\ie, $\rho_L$ = 1.0), a mismatch arises with the conditioning of the denoising model, which prevents achieving optimal performance. Therefore, incorporating a small proportion of the conditional class $\mathbf{c}_{\text{class}}$ is necessary to obtain the best results.

%%%%%%%%%%%%%%%%%%  Further Interpretation LSEP and REPA  %%%%%%%%%%%%%%%%%% 
\subsection{Further Analysis of Combining LSEP and REPA}  \label{appdx:repa_ours}

\begin{figure*}[h]
    \centering
    \includegraphics[width=1.0\linewidth]{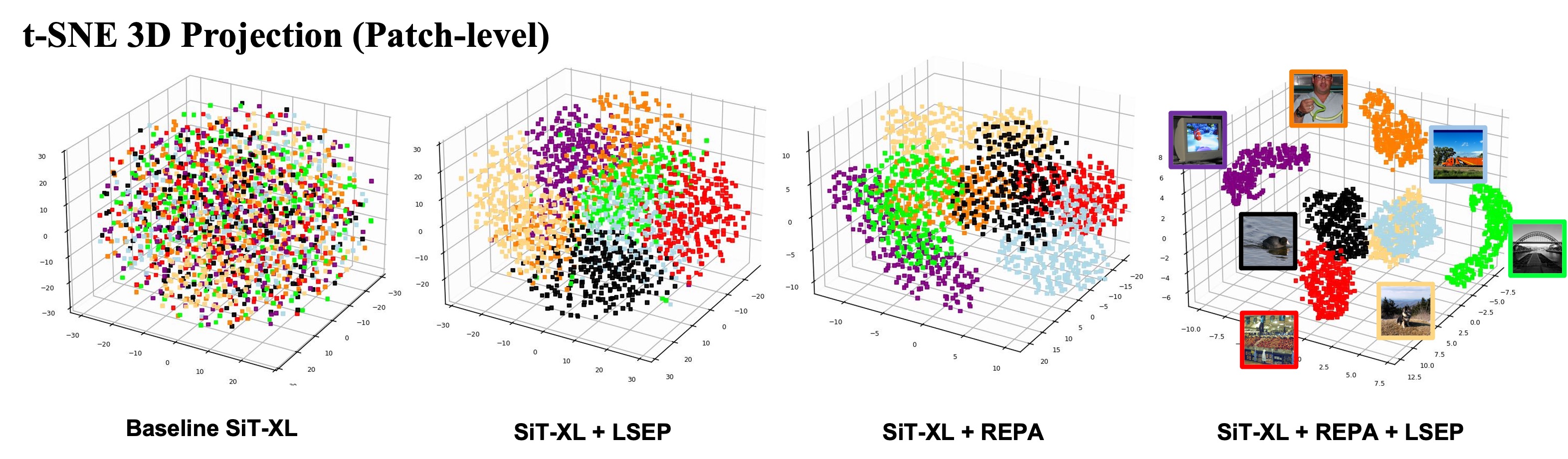} 
    \caption{t-SNE 3D projection visualization of baseline SiT-XL, SiT-XL with our method, SiT-XL with REPA, and SiT-XL with REPA combined with our method in the patch-level manifold space (400K iterations, $t=0.7$), using the same settings as in~\figref{fig:3d_patch_level}.}
   \vspace{-0.3cm} 
    \label{fig:appdx_3d_repa_ours}

\end{figure*}

In~\secref{sec:Quantitative Results}, we showed that combining LSEP with the alignment-based method (REPA) further improves training efficiency and generative performance of the latter. The patch-level linear separability provided by REPA, together with the mean-vector linear separability from LSEP, enhances the learned representations and thereby strengthens training. As shown in~\figref{fig:appdx_3d_repa_ours}, in t-SNE 3D projection, the combination of REPA and LSEP results in the strongest class-wise separability.

\vspace{+0.5cm}
%%%%%%%%%%%%%%%%%%  PCA visualization  %%%%%%%%%%%%%%%%%% 
\subsection{PCA visualization}  \label{appdx:pca_visual}

\begin{figure*}[h]
    \centering
    \includegraphics[width=1.0\linewidth]{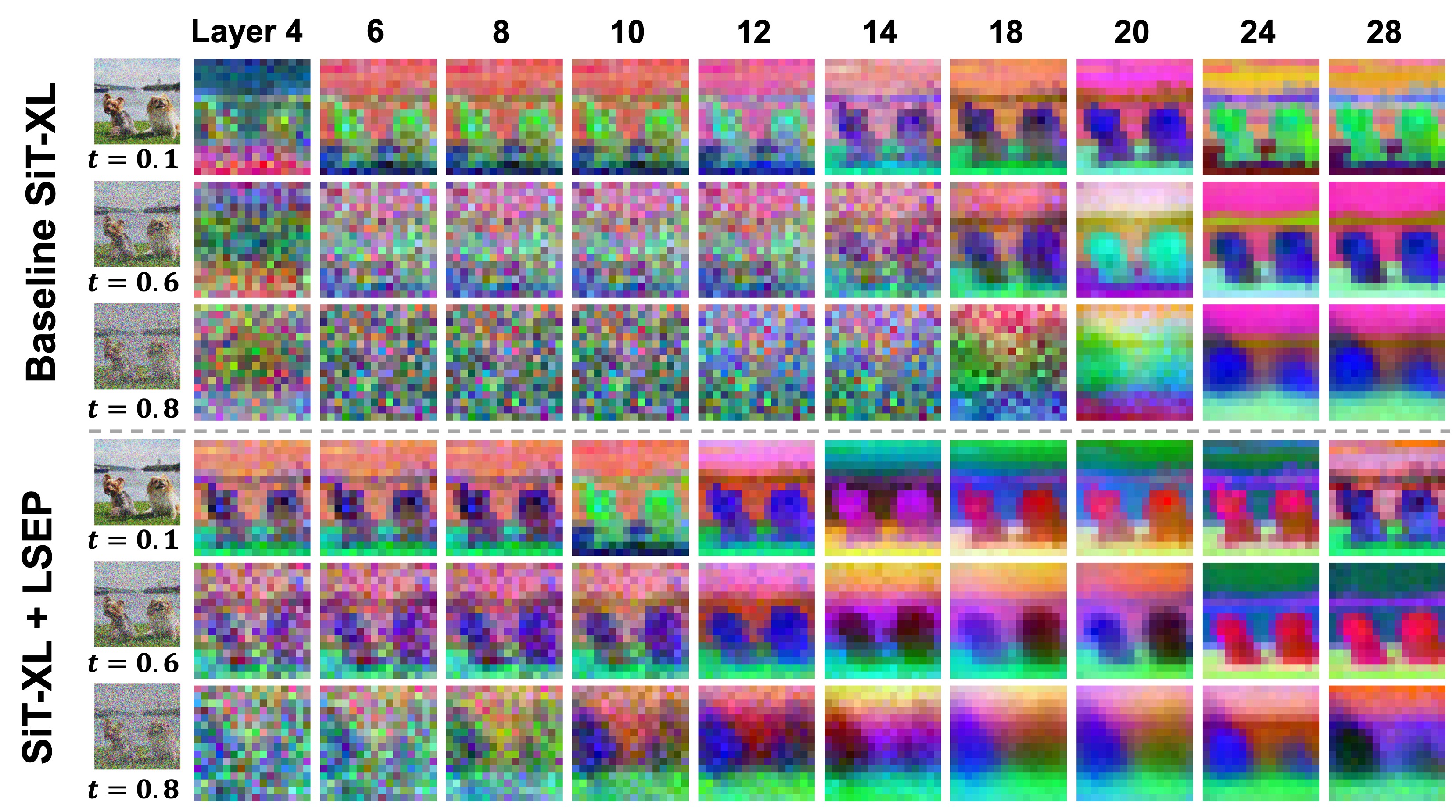} 
    \caption{PCA visualizations at various noise levels and layers within the SiT-XL models. Checkpoints at 400K iterations of baseline SiT-XL and SiT-XL + LSEP are used for the visualizations. }
    \vspace{-0.3cm}
    \label{fig:pca_vis}

\end{figure*}

We plot the PCA visualization of intermediate features. These demonstrate that LSEP preserves distinctly separable components across different noise levels., with particularly clear separation in the early-stage layers and under higher noise conditions (\eg, layers up to 14, $t=0.6$ and $0.8$).

%%%%%%%%%%%%%%%%%%  CFG weights  %%%%%%%%%%%%%%%%%% 
\input{Tables/result_table5}

\section{Detailed evaluation results with the different CFG weights scheduling}  \label{appdx:cfg_guidance}

We provide detailed evaluation results of SiT-XL + LSEP under different classifier-free guidance schedules in~\tabref{tab:LSEP_cfg}. With $\omega_{cfg}$ in the interval [0, 0.65], our LSEP achieves the best performance.

%%%%%%%%%%%%%%%%%%  Additional Qualitative Results %%%%%%%%%%%%%%%%%% 
\section{Additional Qualitative Results}  \label{appdx:more_qual_results}
\begin{figure*}[h]
    \centering
    \includegraphics[width=0.95\linewidth]{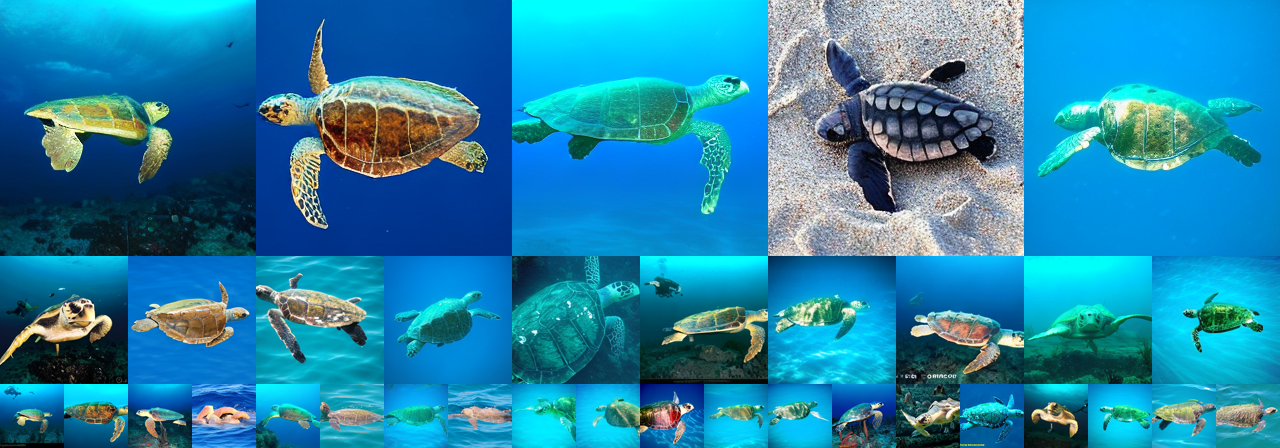} 
    \vspace{-0.3cm}
    \caption{Uncurated samples from SiT-XL/2 + LSEP ($\omega_\text{CFG}=4.0$, class = loggerhead turtle (33))}    
\end{figure*}

\vspace{+0.7cm}

\begin{figure*}[h]
    \centering
    \includegraphics[width=0.97\linewidth]{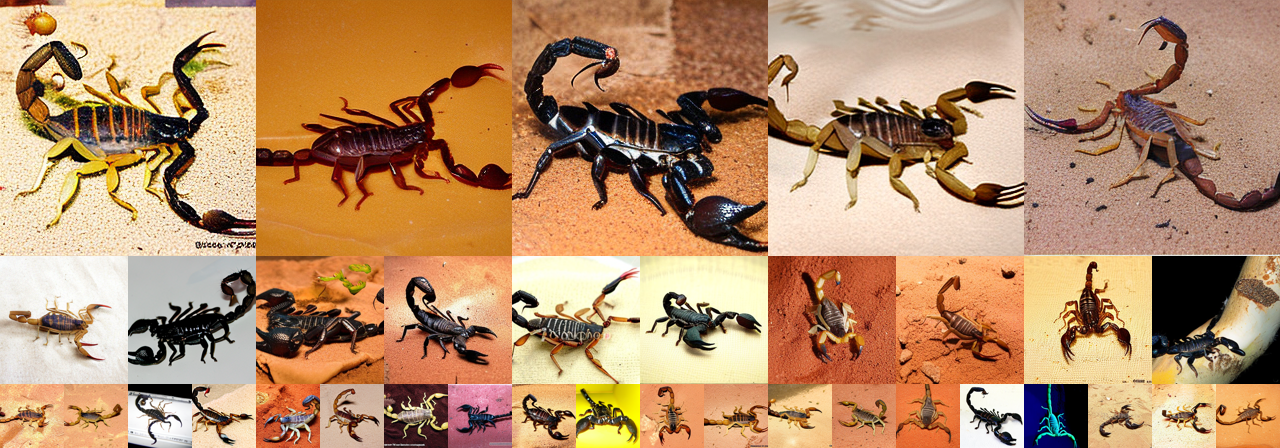} 
    \vspace{-0.3cm}
    \caption{Uncurated samples from SiT-XL/2 + LSEP ($\omega_\text{CFG}=4.0$, class = scorpion (71))}
\end{figure*}

\begin{figure*}[h]
    \centering
    \includegraphics[width=0.97\linewidth]{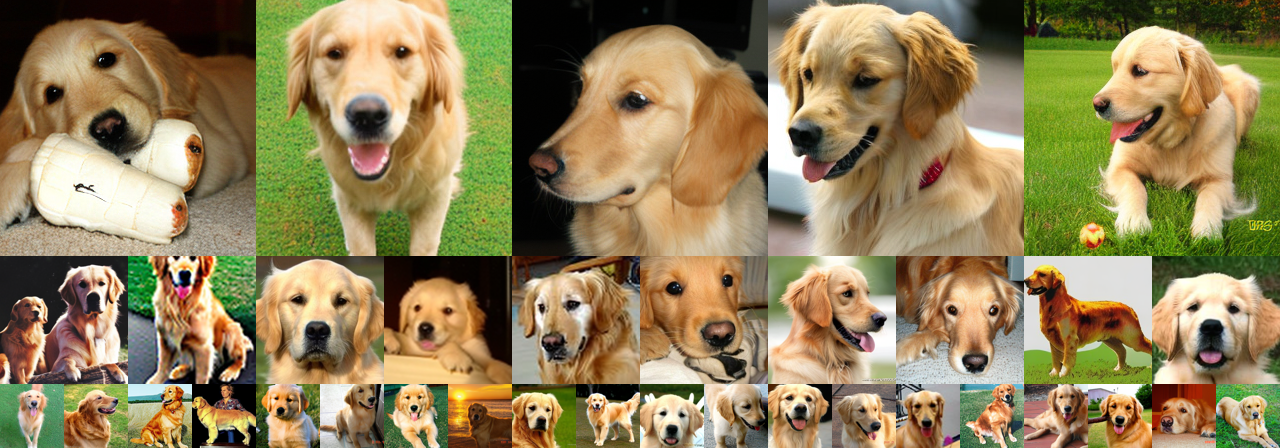} 
    \vspace{-0.3cm}
    \caption{Uncurated samples from SiT-XL/2 + LSEP ($\omega_\text{CFG}=4.0$, class = golden retriever (207))}
\end{figure*}

\begin{figure*}[h]
    \centering
    \includegraphics[width=0.97\linewidth]{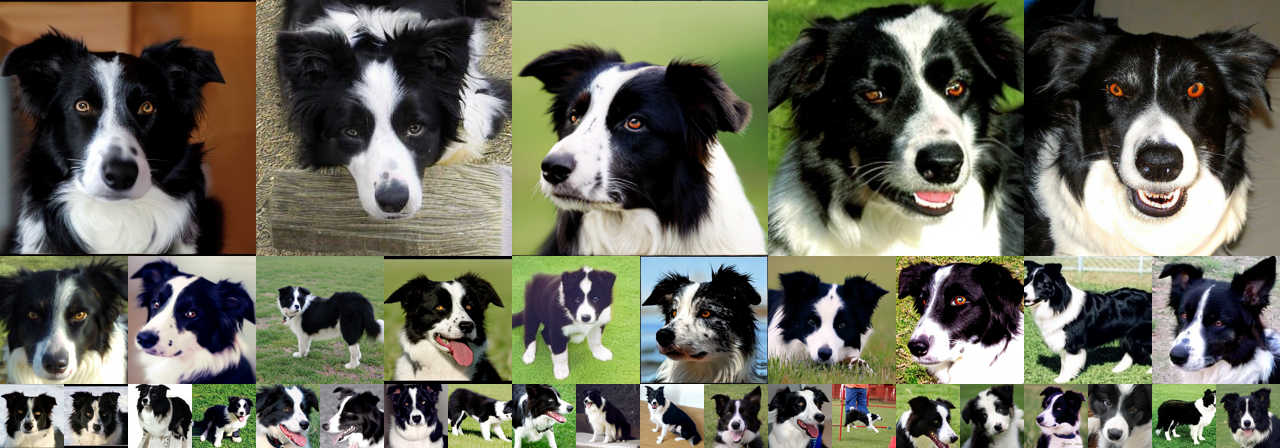} 
    \vspace{-0.3cm}
    \caption{Uncurated samples from of SiT-XL/2 + LSEP ($\omega_\text{CFG}=4.0$, class = border collie(232))}
\end{figure*}

\begin{figure*}[h]
    \centering
    \includegraphics[width=0.97\linewidth]{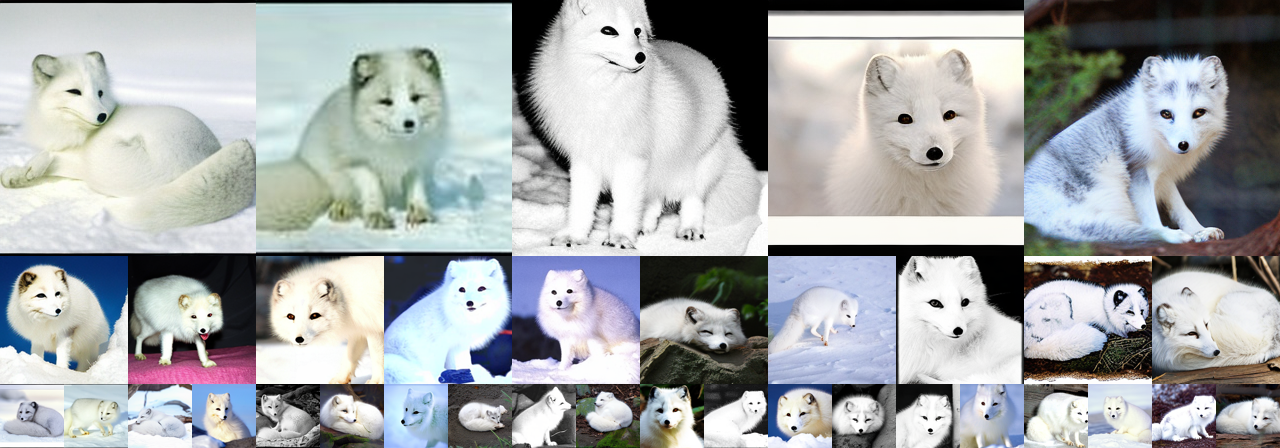} 
    \vspace{-0.3cm}
    \caption{Uncurated samples from SiT-XL/2 + LSEP ($\omega_\text{CFG}=4.0$, class = arctic fox (279))}
\end{figure*}

\begin{figure*}[h]
    \centering
    \includegraphics[width=0.97\linewidth]{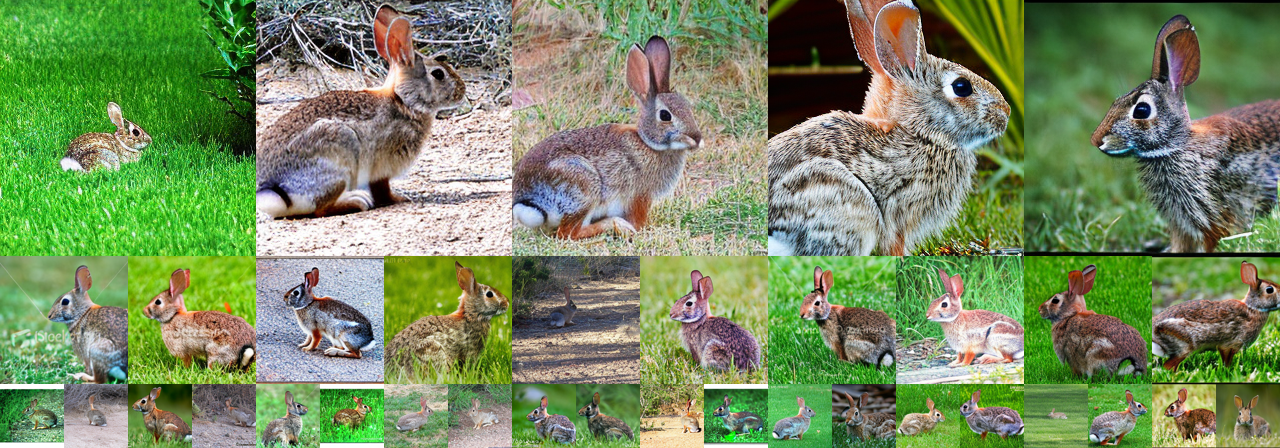} 
    \vspace{-0.3cm}
    \caption{Uncurated samples from SiT-XL/2 + LSEP ($\omega_\text{CFG}=4.0$, class = wood rabbit (330))}

\end{figure*}

\begin{figure*}[h]
    \centering
    \includegraphics[width=0.97\linewidth]{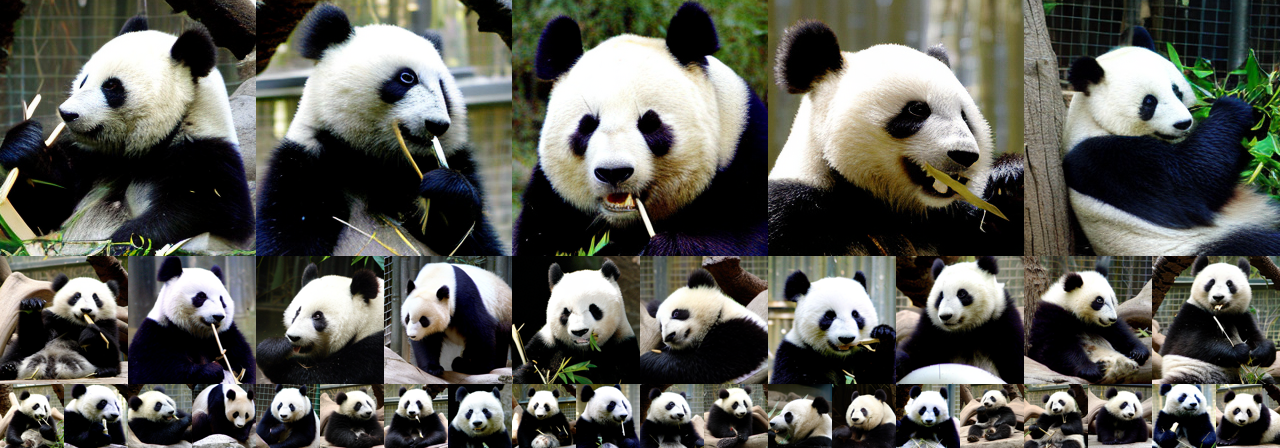} 
    \vspace{-0.3cm}
    \caption{Uncurated samples from SiT-XL/2 + LSEP ($\omega_\text{CFG}=4.0$, class = panda (388))}
\end{figure*}

\begin{figure*}[h]
    \centering
    \includegraphics[width=0.97\linewidth]{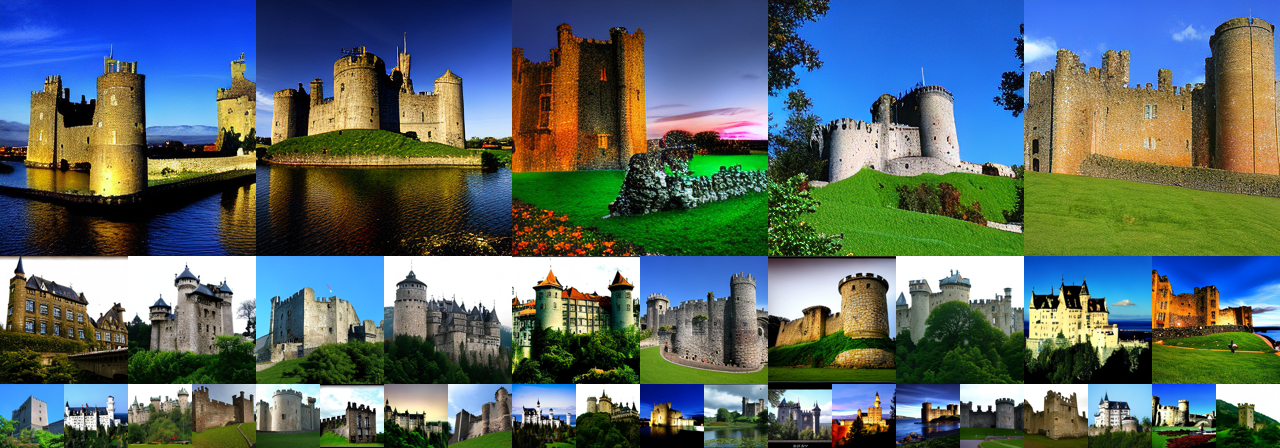} 
    \vspace{-0.3cm}
    \caption{Uncurated samples from SiT-XL/2 + LSEP ($\omega_\text{CFG}=4.0$, class = castle (483))}
\end{figure*}

\begin{figure*}[h]
    \centering
    \includegraphics[width=0.97\linewidth]{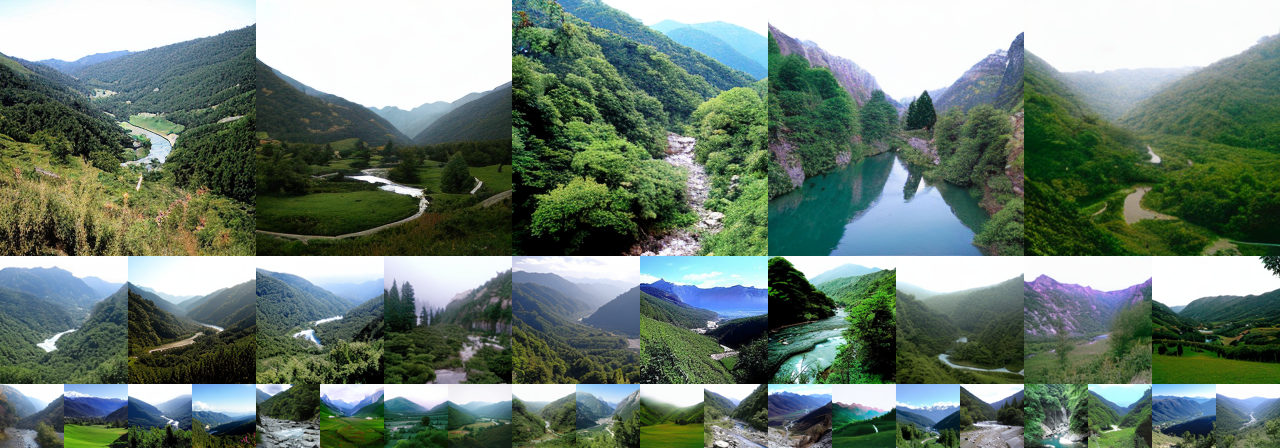} 
    \vspace{-0.3cm}
    \caption{Uncurated samples from SiT-XL/2 + LSEP ($\omega_\text{CFG}=4.0$, class = valley (797))}
\end{figure*}

\begin{figure*}[h]
    \centering
    \includegraphics[width=0.97\linewidth]{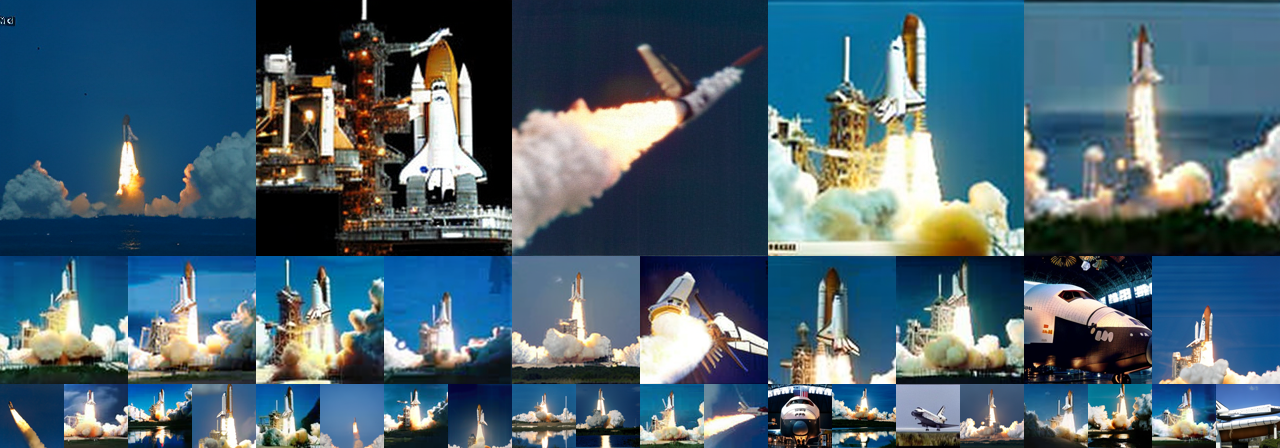} 
    \vspace{-0.3cm}
    \caption{Uncurated samples from SiT-XL/2 + LSEP ($\omega_\text{CFG}=4.0$, class = space shuttle (812))}
\end{figure*}

\end{document}

%% file: mics/math_commands.tex
\usepackage{amsmath,amsfonts,bm}

\def\figref#1{figure~\ref{#1}}
\def\secref#1{section~\ref{#1}}
\def\eqref#1{equation~\ref{#1}}
\def\1{\bm{1}}

\DeclareMathAlphabet{\mathsfit}{\encodingdefault}{\sfdefault}{m}{sl}
\SetMathAlphabet{\mathsfit}{bold}{\encodingdefault}{\sfdefault}{bx}{n}

%% file: Tables/result_table1.tex
\begin{table*}[t]
\vskip -0.5in
\caption{Performance comparison of the proposed key training strategies on the SiT-L/2 model. Each strategy is distinguished by a different color, and dark colors indicate the best option within each component. The symbols $\uparrow$ and $\downarrow$ indicate that higher and lower values are better, respectively.}
\vspace{-0.3cm}
\label{tab:cond_lp_results}
\begin{center}
\renewcommand{\arraystretch}{1.05}
\setlength{\tabcolsep}{2.3pt}
\begin{small}
% \begin{sc}
\begin{tabular}{ccccccccccc}
\hline
\rowcolor{black!10}
\textbf{Iter.} & \textbf{Uncond. Prob.} & \textbf{Target} & \textbf{Rand. Crop ($n \! \times \! n$)} & \textbf{$\omega_{\text{class}}$} & \textbf{FID$\downarrow$} & \textbf{sFID$\downarrow$} & \textbf{IS$\uparrow$} & \textbf{Pre$\uparrow$} & \textbf{Rec$\uparrow$} \\
\hline
\rowcolor{yellow!10}
400K & \multicolumn{4}{c}{\textbf{Baseline SiT-L/2}}  & 18.5 & 5.20 & 72.2 & 0.64 & 0.63 \\
% & & SiT-L/2 + REPA & -- & 10.00 & 5.23 & 111.21 & 0.68 & 0.66 \\
\arrayrulecolor{gray} \cmidrule(lr){1-10}

% \rowcolor{cyan!10}
% \multicolumn{10}{l}{\textbf{SiT-L/2 + Ours}}  \\
% \arrayrulecolor{gray} \cmidrule(lr){1-10}

%%%% Uncondition Probability %%%%
\multirow{4}{*}{400K} 
& \cellcolor{blue!10} 0.1 & 8  & \multirow{4}{*}{$n=16$} & \multirow{4}{*}{0.03} & 20.6 & 5.47 & 70.2& 0.62 & 0.65 \\
& \cellcolor{blue!10} 0.8 & 8  &  & & 13.1 & 5.33 & 95.2 & 0.67 & 0.63 \\
& \cellcolor{blue!20} \textbf{0.9} & 8 &  & & 12.9 & 5.37 & 96.2 & 0.67 & 0.64 \\
& \cellcolor{blue!10} 1.0 & 8 &  & & 14.1 & 5.34 & 91.0 & 0.66 & 0.64 \\
\arrayrulecolor{gray} \cmidrule(lr){1-10}

%%%% Target Depth %%%%
\multirow{5}{*}{400K} & \multirow{5}{*}{0.9} 
& \cellcolor{orange!10} 6  & \multirow{5}{*}{$n=16$} & \multirow{5}{*}{0.03} & 12.8 & 5.41 & 96.2 & 0.67 & 0.64 \\
& & \cellcolor{orange!30} \textbf{7}  & & & 12.7 & 5.34 & 98.1 & 0.67 & 0.64 \\
& & \cellcolor{orange!10} 8  & & & 12.9 & 5.37 & 96.2 & 0.67 & 0.64 \\
& & \cellcolor{orange!10} 9  & & & 13.9 & 5.37 & 91.5 & 0.66 & 0.65 \\
& & \cellcolor{orange!10} 10  & & & 13.4 & 5.43 & 93.0 & 0.66 & 0.64 \\
\arrayrulecolor{gray} \cmidrule(lr){1-10}

%%%% Random Crop %%%%
\multirow{5}{*}{400K} & \multirow{5}{*}{0.9} 
& 7  & $n=16$ & 0.03 & 12.7 & 5.34 & 98.1 & 0.67 & 0.64 \\
& & 7 & \cellcolor{green!10} $n \in [11, 16] \cap \mathbb{Z} $ & 0.03 & 13.0 & 5.33 & 95.8 & 0.67 & 0.64 \\
& & 7 & \cellcolor{green!30} $\mathbf{n} \in [\mathbf{12}, \mathbf{16}] \cap \mathbb{Z} $ & 0.03 & 12.5 & 5.34 & 98.8 & 0.67 & 0.64 \\
& & 7 & \cellcolor{green!10} $n \in [13, 16] \cap \mathbb{Z} $ & 0.03 & 13.2 & 5.33 & 95.2 & 0.67 & 0.64 \\
& & 7 & \cellcolor{green!10} $n \in [14, 16] \cap \mathbb{Z} $ & 0.03 & 12.5 & 5.37 & 97.6 & 0.68 & 0.64 \\
\arrayrulecolor{gray} \cmidrule(lr){1-10}

%%%% Weighting Scheduling %%%%
\multirow{6}{*}{400K} & \multirow{6}{*}{0.9} 
& 7 & $n=16$ & 0.03 & 12.7 & 5.34 & 98.1 & 0.67 & 0.64 \\

& & 7 & $n=16$ & \cellcolor{cyan!10}  $[0.02, 0.03]_{\text{10 Bins}}$ & 12.5 & 5.33 & 97.4 & 0.67 & 0.64 \\

& & 7 & $n=16$ & \cellcolor{cyan!10}  $[0.02, 0.03]_{\text{5 Bins}}$ & 12.5 & 5.36 & 98.6 & 0.67 & 0.64 \\

% & & 7 & $n=16$ & \cellcolor{cyan!10}  $[0.025, 0.03]_{\text{10 Bins}}$ & 12.6 & 5.35 & 98.1 & 0.67 & 0.64 \\

& & 7 & $n \in [12, 16] \cap \mathbb{Z} $ & \cellcolor{cyan!10}  $[0.02, 0.025]_{\text{10 Bins}}$ & 12.5 & 5.32 & 97.9 & 0.67 & 0.63 \\

& & 7 & $n \in [12, 16] \cap \mathbb{Z} $ & \cellcolor{cyan!10}  $[0.0275, 0.0325]_{\text{5 Bins}}$ & 12.6 & 5.41 & 97.7 & 0.67 & 0.64 \\

& & 7 & $n \in [12, 16] \cap \mathbb{Z} $ & \cellcolor{cyan!30}  $[\mathbf {0.0275}, \mathbf {0.0325}]_{\text{10 Bins}}$ & 12.3 & 5.40 & 98.3 & 0.68 & 0.64 \\

\arrayrulecolor{black} \hline

\end{tabular}
% \end{sc}
\end{small}
\end{center}
\label{tab:table_results}
\vskip -0.1in
\end{table*}

%% file: Tables/result_table2.tex
\begin{table}[H]
\centering
    \vspace{-0.2cm}  
    %\caption{FID comparisons between our method and standard SiTs on $256 \times 256$ ImageNet, without employing CFG. %The symbol $\downarrow$ indicates that lower values are better.
    \caption{FID comparisons on $256 \times 256$ ImageNet, without employing CFG.
    $\downarrow$ indicates lower values are better.}
    \vspace{+0.1cm}  
    \label{tab:fid_comparison}
    \begin{scriptsize} %scriptsize
    \begin{tabular}{lccc}
    \toprule
    \textbf{Model} & \textbf{\#Params} & \textbf{Iter.} & \textbf{FID$\downarrow$} \\
    \midrule
    SiT-B/2 & 130M & 400K & 33.0 \\
    SiT-B/2 + REPA & 137M & 400K & 24.4 \\
    SiT-B/2 + \textbf{LSEP (ours)} & 131M & 400K & 28.3 \\
    SiT-B/2 + REPA + \textbf{LSEP (ours)} & 139M & 400K & \textbf{20.5} \\
    \midrule
    SiT-L/2 & 458M & 400K & 18.8 \\
    SiT-L/2 + REPA & 466M & 400K & 9.7 \\
    SiT-L/2 + \textbf{LSEP (ours)} & 459M & 400K & 12.3 \\
    SiT-L/2 + REPA + \textbf{LSEP (ours)} & 467M & 400K & \textbf{9.5} \\
    
    \midrule
    SiT-XL/2 & 675M & 400K & 17.2 \\
    SiT-XL/2 + REPA & 683M & 400K & 7.9 \\
    SiT-XL/2 + \textbf{LSEP (ours)} & 676M & 400K & 10.4 \\
    SiT-XL/2 + REPA + \textbf{LSEP (ours)} & 684M & 400K & \textbf{7.5} \\
    % \arrayrulecolor{gray} \cmidrule(lr){1-4}
    
    % SiT-XL/2 & 675M & 7M & 8.3 \\
    % SiT-XL/2 + REPA & 683M & 4M & 5.9 \\
    % SiT-XL/2 + \textbf{LSEP (ours)} & 676M & 2M & 6.8 \\
    % SiT-XL/2 + \textbf{LSEP (ours)} & 676M & 4M & 00.0 \\
        
    \bottomrule
\label{tab:table_results2}
\end{tabular}
\end{scriptsize}
\end{table}

%% file: Tables/result_table3.tex
% \begin{SCtable}[][t] 
\begin{table}[t]

\renewcommand{\arraystretch}{1.1}

\centering
\caption{\textbf{System-level comparison} on ImageNet $256 \times 256$ with CFG. 
Within each architecture, the \textbf{best} on each metric are bolded. Arrows indicate whether lower $(\downarrow)$ or higher $(\uparrow)$ values are better.}
\vspace{+0.1cm}
\label{tab:system_level}
\begin{small}

\resizebox{\textwidth}{!}{%
\begin{tabular}{lccccccc}

\toprule
\rowcolor{black!10} \textbf{Model} & \textbf{Epochs} & \textbf{Tokenizer} & \textbf{FID}$\downarrow$ & \textbf{sFID}$\downarrow$ & \textbf{IS}$\uparrow$ & \textbf{Pre.}$\uparrow$ & \textbf{Rec.}$\uparrow$ \\
\midrule

\multicolumn{8}{l}{\textbf{Pixel diffusion}} \\
ADM-U~\citep{dhariwal2021beatGANs} & 400 & - & 3.94 & \textbf{6.14} & 186.7 & \textbf{0.82} & \textbf{0.52} \\
VDM++~\citep{kingma2023understanding} & 560 & - & \textbf{2.40} & - & \textbf{225.3} & - & - \\
Simple diffusion~\citep{hoogeboom2023simple} & 800 & - & 2.77 & - & 211.8 & - & - \\
CDM~\citep{ho2022cascaded} & 2160 & - & 4.88 & - & 158.7 & - & - \\
\arrayrulecolor{gray} \cmidrule(lr){1-8}

\multicolumn{8}{l}{\textbf{Latent diffusion, U-Net}} \\
LDM-4~\citep{rombach2022StableDiffusion} & 200 & LDM-VAE & \textbf{3.60} & - & \textbf{247.7} & \textbf{0.87} & \textbf{0.48} \\
\arrayrulecolor{gray} \cmidrule(lr){1-8}

\multicolumn{8}{l}{\textbf{Latent diffusion, Transformer \emph{with} pre-trained external encoder}} \\
SiT + REPA~\citep{yu2025REPA} & 800 & SD-VAE & 1.42 & 4.70 & 305.7 & \textbf{0.80} & 0.65 \\
LightningDiT~\citep{yao2025reconstruction} & 800 & VA-VAE & 1.35 & 4.15 & 295.3 & 0.79 & 0.65 \\
REPA-E~\citep{leng2025REPAE} & 800 & VA-VAE & \textbf{1.26} & \textbf{4.11} & \textbf{314.9} & 0.79 & \textbf{0.66} \\
\arrayrulecolor{gray} \cmidrule(lr){1-8}

\multicolumn{8}{l}{\textbf{Latent diffusion, Transformer \emph{without} pre-trained external encoder}} \\
DiT-XL/2~\citep{peebles2023scalable} & 1400 & SD-VAE & 2.27 & 4.60 & 278.2 & 0.83 & 0.57 \\
SiT-XL/2~\citep{ma2024sit} & 1400 & SD-VAE & 2.06 & \textbf{4.50} & 270.3 & 0.82 & 0.59 \\
SD-DiT~\citep{zhu2024sd} & 480 & SD-VAE & 3.23 & - & - & - & - \\
MaskDiT~\citep{Zheng2024MaskDiT} & 1600 & SD-VAE & 2.28 & 5.67 & 276.6 & 0.80 & 0.61 \\
DiT + TREAD~\citep{krause2025tread} & 740 & SD-VAE & 1.69 & 4.73 & 292.7 & \textbf{0.81} & 0.63 \\
% SiT + MAETok & 800 & MAE-Tok & 1.67 & - & 311.2 & - & - \\
SiT + SRA~\citep{jiang2025SREPA} & 800 & SD-VAE & 1.58 & 4.65 & \textbf{311.4} & 0.80 & 0.63 \\
\rowcolor{brown!30} SiT + LSEP (ours) & 800 & SD-VAE & \textbf{1.46} & 4.94 & 296.8 & 0.80 & \textbf{0.64} \\

\bottomrule
\end{tabular}
}
\end{small}
\end{table}
% \end{SCtable}

%% file: Tables/table_hyperparameters.tex
\begin{table}[h]
\centering
\caption{Hyperparameter setup for different SiT models and LSEP.}
\label{tab:hyperparams}
\begin{small}
\begin{tabular}{lccc}
\toprule
 & \textbf{SiT-B/2} & \textbf{SiT-L/2} & \textbf{SiT-XL/2} \\
\midrule
\textbf{Architecture} & & & \\
Input dim. & $32\times32\times4$ & $32\times32\times4$ & $32\times32\times4$ \\
Num. layers & 12 & 24 & 28 \\
Hidden dim. & 768 & 1,024 & 1,152 \\
Num. heads & 12 & 16 & 16 \\
\midrule
\textbf{LSEP} & & & \\
Uncon. Prob. & 90\% & 90\% & 90\% \\
Target depth & 4 & 7 & 8 \\
Rand. Crop & [14, 16] & [12, 16] & [12, 16] \\
$\omega(t)$ & $[\text{0.005}, \text{0.01}]_{10 \ \mathrm{bins}}$ & $[\text{0.0275}, \text{0.0325}]_{10 \ \mathrm{bins}}$ & $[\text{0.0225}, \text{0.03}]_{10 \ \mathrm{bins}}$ \\
lr for linear probes. & 0.03 & 0.0001 & 0.0001 \\

\midrule
\textbf{Optimization} & & & \\
Training iteration & 400K & 400K & 400K / 4M (Fig. 1) \\
Batch size & 256 & 256 & 256 \\
Optimizer & AdamW & AdamW & AdamW \\
lr & 0.0001 & 0.0001 & 0.0001 \\
$(\beta_1,\beta_2)$ & (0.9,0.999) & (0.9,0.999) & (0.9,0.999) \\
\midrule
\textbf{Interpolants} & & & \\
$\alpha_t$ & $1-t$ & $1-t$ & $1-t$ \\
$\sigma_t$ & $t$ & $t$ & $t$ \\
$w_t$ & $\sigma_t$ & $\sigma_t$ & $\sigma_t$ \\
Training objective & v-prediction & v-prediction & v-prediction \\
Sampler & Euler-Maruyama & Euler-Maruyama & Euler-Maruyama \\
Sampling steps & 250 & 250 & 250 \\
Guidance & - & - & $\omega_{cfg}=1.8, \text{Interval : }[0, 0.65]$ (Tab.3) \\

\bottomrule
\label{tab:table_hyperparameters}
\end{tabular}
\end{small}
\end{table}

%% file: Tables/result_table5.tex
\begin{table}[h]
\centering
\vspace{+0.5cm}
\caption{Evaluation results of SiT-XL + LSEP at 4M iteration with different CFG weight and guidance interval.\vspace{6pt}}
\label{tab:LSEP_cfg}
\begin{scriptsize}
\begin{tabular}{lcc|cc|ccccc}
\toprule
Model & \#Params & Iter. & $w_{cfg}$ & Interval & FID$\downarrow$ & sFID$\downarrow$ & IS$\uparrow$ & Pre.$\uparrow$ & Rec.$\uparrow$ \\
\midrule
\rowcolor{yellow!10} SiT-XL/2~\citep{ma2024sit}  & 675M & 7M & [0, 1]   & 1.50 & 2.06 & 4.50 & 270.3 & 0.82 & 0.59 \\
\midrule
+ LSEP (ours) & 675M & 4M & 1.95 & [0, 0.75] & 2.18 & 4.55 & 344.3 & 0.83 & 0.60 \\
+ LSEP (ours) & 675M & 4M & 1.95 &  [0, 0.70] & 1.81 & 4.68 & 328.6 & 0.82 & 0.62 \\
+ LSEP (ours) & 675M & 4M & 1.95 & [0, 0.65] & 1.59 & 4.87 & 312.2 & 0.81 & 0.63 \\
+ LSEP (ours) & 675M & 4M & 1.95 & [0, 0.60] & 1.49 & 5.15 & 296.7 & 0.79 & 0.64 \\
+ LSEP (ours) & 675M & 4M & 1.90 & [0, 0.60] & 1.50 & 5.15 & 291.8 & 0.79 & 0.64 \\
+ LSEP (ours) & 675M & 4M & 1.85 & [0, 0.60] & 1.56 & 5.10 & 282.6 & 0.79 & 0.65 \\
+ LSEP (ours) & 675M & 4M & \textbf{1.80} & \textbf{[0, 0.65]} & 1.46 & 4.94 & 296.8 & 0.80 & 0.64 \\
+ LSEP (ours) & 675M & 4M & 1.80 & [0, 0.60] & 1.55 & 5.17 & 280.6 & 0.79 & 0.65 \\

% + LSEP (ours) & 675M & 4M & [0, 0.65] & 1.80 & a27 in prog. & 0.00 & 0.00 & 0.0 & 0.00 \\

\bottomrule
\end{tabular}
\end{scriptsize}
\end{table}